%% file: main.tex
\documentclass[10pt,twocolumn,letterpaper]{article}

\usepackage[pagenumbers]{cvpr} 

\usepackage{graphicx}
\usepackage{amsmath}
\usepackage{amssymb}
\usepackage{booktabs}
\usepackage{multirow}
\usepackage{xcolor}
\usepackage{dsfont}
\usepackage[inline]{enumitem}
\usepackage{makecell}
\usepackage{soul}
\usepackage{animate}
\usepackage{titling}

\usepackage[pagebackref,breaklinks,colorlinks]{hyperref}

\usepackage[capitalize]{cleveref}
\crefname{section}{Sec.}{Secs.}
\Crefname{section}{Section}{Sections}
\Crefname{table}{Table}{Tables}
\crefname{table}{Tab.}{Tabs.}

\def\cvprPaperID{12007} 
\def\confName{CVPR}
\def\confYear{2023}

\begin{document}

\title{\Large \bf Temporally Consistent Online Depth Estimation Using Point-Based Fusion}

\author{Numair Khan\\
{\tt\small numairkhan@meta.com}
\and
Eric Penner\\
{\tt\small epenner@meta.com}
\and
Douglas Lanman\\
{\tt\small douglas.lanman@meta.com}
\and
Lei Xiao\\
{\tt\small lei.xiao@meta.com}
}
\date{\vspace{-3mm}}
\maketitle

\input{macros}
\input{src/0.abstract}
\input{src/1.introduction}

\input{src/2.related-work}
\input{src/3.method}

\input{src/3a.temporal-filtering}

\input{src/3b.depth-fusion}

\input{src/3c.global-pc-update}

\input{src/3d.training}

\input{src/4.experiments}

\input{src/5.conclusion}

\section*{Acknowledgements}
We thank Julia Majors and Thu Nguyen-Phuoc for proofreading the draft. Zhao Dong and Zhaoyang Lv provided useful insights on the technical aspects of the project.

{\small
\bibliographystyle{ieee_fullname}
\bibliography{bibliography}
}

\clearpage
\title{\Large \bf \vspace{-6mm}Supplemental Document for\\ Temporally Consistent Online Depth Estimation Using Point-Based Fusion\vspace{-8mm}}
\author{}
\date{}
\maketitle

\setcounter{table}{0}
\renewcommand{\thetable}{A\arabic{table}}
\renewcommand{\thesection}{A\arabic{section}}
\renewcommand{\thefigure}{A\arabic{figure}}

\input{src/A1.results}

\input{src/A2.architecture}

\end{document}

%% file: macros.tex
\newcommand{\mat}[1]{\mathrm{\textbf{#1}}}
\newcommand{\vect}[1]{\mathrm{\textbf{#1}}}

\definecolor{mygreen}{rgb}{0, 0.8, 0}

\newcommand{\obv}[1]{#1^t}
\newcommand{\prior}[1]{#1^t_\mathrm{p}} 
\newcommand{\dtempt}[1]{d^{#1}_\mathrm{f}}
\newcommand{\dtemp}{\dtempt{t}} 
\newcommand{\dout}{d^t_\mathrm{o}} 

%% file: src/0.abstract.tex
\begin{abstract}
Depth estimation is an important step in many computer vision problems such as 3D reconstruction, novel view synthesis, and computational photography. Most existing work focuses on depth estimation from single frames. When applied to videos, the result lacks temporal consistency, showing flickering and swimming artifacts. In this paper we aim to estimate temporally consistent depth maps of video streams in an online setting. This is a difficult problem as future frames are not available and the method must choose between enforcing consistency and correcting errors from previous estimations. The presence of dynamic objects further complicates the problem. We propose to address these challenges by using a global point cloud that is dynamically updated each frame, along with a learned fusion approach in image space. Our approach encourages consistency while simultaneously allowing updates to handle errors and dynamic objects. Qualitative and quantitative results show that our method achieves state-of-the-art quality for consistent video depth estimation.
\end{abstract}

%% file: src/1.introduction.tex
\section{Introduction}
\label{sec:introduction}
\input{figures/teaser}
Depth reconstruction is a long-standing, fundamental problem in computer vision. For decades the most popular depth estimation techniques were based on stereo matching~\cite{schoenberger2016mvs} or structure-from-motion~\cite{schoenberger2016sfm}. However, more recently the best results have come from learning-based approaches~\cite{middlebury2022}. As the overall reconstruction quality has improved, focus has shifted to new areas such as monocular estimation~\cite{li2019, ranftl2020, ranftl2021}, edge quality~\cite{zhu2020}, and temporal consistency~\cite{luo2020, li2021}. The latter is particularly important for video applications in computational photography and virtual reality~\cite{chaurasia2020, xiao2022} as inconsistent depth can cause objectionable flickering and swimming artifacts. 

Consistent video depth estimation, however, remains a difficult problem as even the best method will suffer from unpredictable errors and imperfections based on scene content, especially in textureless and specular regions. This difficulty is aggravated by many of the aforementioned applications requiring online reconstruction: future frames are not known beforehand and temporal consistency must be balanced with error-correction. Furthermore, the presence of dynamic objects --- which are inherently inconsistent --- adds an additional layer of complication. Luo~\etal~\cite{luo2020} address these problems by assuming all frames are known beforehand and fine-tuning their method for each input video. Other approaches seek an online solution by encoding consistency in network weights either through a training loss~\cite{li2019}, by using recurrent architectures~\cite{kumar2018, patil2020, zhang2019}, or by conditioning it on the input~\cite{liu2019}. Each method, however, ultimately relies on the raw output of a neural network being consistent, which is difficult to achieve due to camera noise and aliasing in convolutional networks~\cite{karras2021, vasconcelos2021, zhang2019shift}.

\input{figures/pipeline}

In this work, we propose the use of a global point cloud to encourage temporal consistency in online video depth estimation. We demonstrate how to tackle the twin problems of handling dynamic objects and updating a static --- and potentially erroneous --- point cloud when future frames are not known. With quantitative and qualitative results, we show our approach significantly improves the temporal consistency of both stereo and monocular depth estimation without sacrificing spatial quality. In summary, our contributions are as follows:
\begin{enumerate}[topsep=2pt,itemsep=-0.5ex]
    \item We propose point cloud-based fusion for temporally consistent video depth estimation.
    \item We present a three-stage approach to encourage consistency in online settings, while simultaneously allowing updates to improve the accuracy of reconstruction and handle dynamic scenes. 
    \item We present an image-space approach to dynamics estimation and depth fusion that is lightweight and has low runtime overhead. 
\end{enumerate}

%% file: figures/teaser.tex
\begin{figure}[t]
  \centering
   \includegraphics[width=1.0\linewidth, trim={6.5cm, 0cm, 0cm, 0cm}, clip]{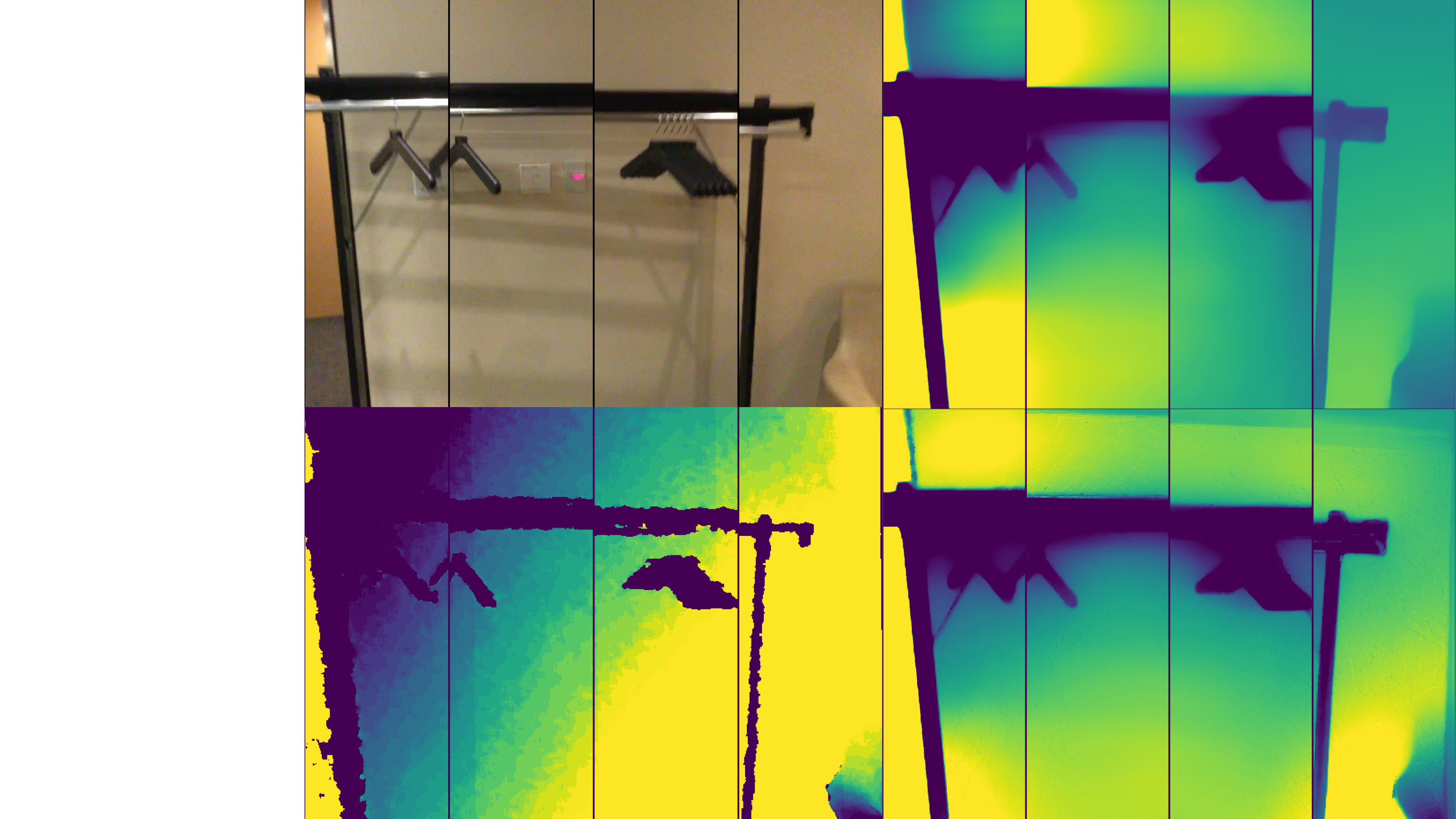}
   \caption{Comparing monocular depth estimation on the ScanNet dataset across four frames. \textit{Clockwise from Top-left:}  Input RGB image, Ranftl~\etal's DPT~\cite{ranftl2021},  our result with a DPT backbone, RGB-D sensor ground truth. }
   \label{fig:teaser}
   \vspace{-4mm}
\end{figure}

%% file: figures/pipeline.tex
\begin{figure*}[t!]
  \centering
   \includegraphics[width=0.85\linewidth, trim={6.7cm, 5.8cm, 7.1cm, 8.2cm}, clip]{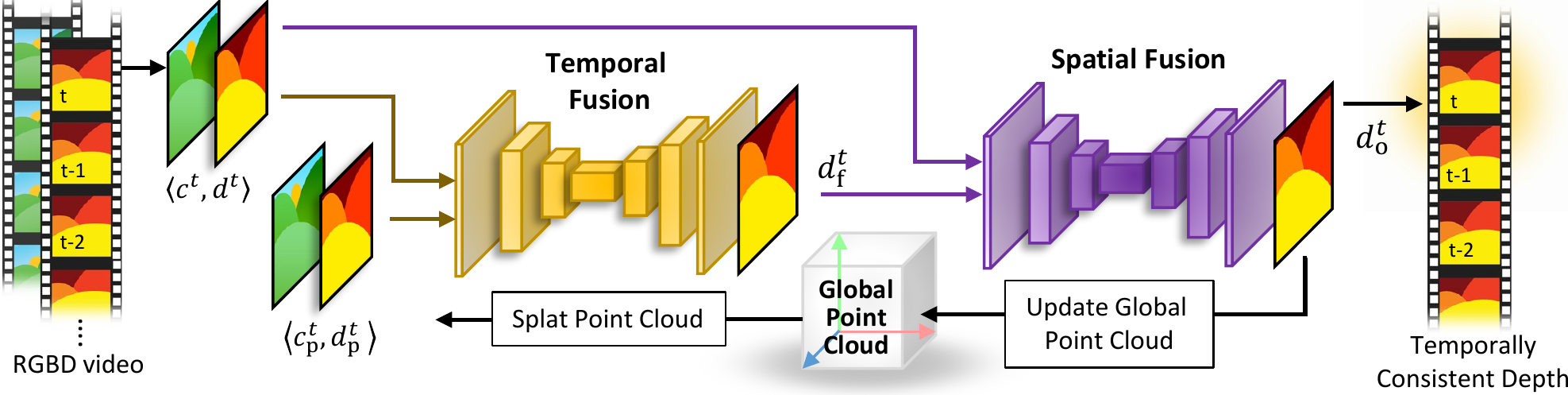}
   \caption{We generate temporally consistent depth maps for each RGB video frame $t$ by fusing the projected depth from a prior point cloud $\prior{d}$ with the estimated depth $d^t$. This is done by temporally fusing $\prior{d}$ to update dynamic regions, followed by spatial fusion with $d^t$ based on confidence estimates of accuracy. The final result $\dout$ is used to update the point cloud for the next frame. }
   \label{fig:pipeline}
   \vspace{-4mm}
\end{figure*}

%% file: src/2.related-work.tex
\section{Related Work}
\label{sec:related-work}
\textbf{Depth Estimation:} The recovery of geometry from color images is an historically important problem in computer vision. In the last decade, deep learning has increasingly been applied to different stages of classic algorithms including geometric~\cite{lipson2021, tankovich2021, xu2022, cheng2020}, photometric~\cite{xu2019} and multi-view stereo~\cite{huang2018}, as well as structure-from-motion (SfM)~\cite{wei2020}. More recently, approaches based entirely on learnt priors have allowed high-quality depth estimation from single images~\cite{ranftl2020, ranftl2021, wang2019, li2019}. Traditionally, the focus of these methods has been on single frames, with most suffering from temporal artifacts on videos. 

\textbf{Temporal Consistency:} A number of works have recently demonstrated the advantage of using temporal information for improving depth accuracy and consistency~\cite{liu2019, long2021, patil2020}. 
Long~\etal~\cite{long2021} propose temporal coherence by fusing the cost volumes of neighboring frames with a transformer. Similarly, Liu~\etal's~\cite{liu2019} method accumulates depth probability over consecutive frames using Bayesian filtering. 
However, the final outcome in both cases is generated by a refinement network which does not have any explicit temporal constraints. 
Luo~\etal~\cite{luo2020} fine-tune a monocular depth network at inference-time using constraints from optical flow and structure-from-motion. Kopf~\etal\cite{kopf2021} build on this to refine camera poses too. Both methods run offline with the inference-time fine-tuning taking around twenty minutes.

Luo~\etal and Kopf~\etal are special instances of a more general approach to enforcing consistency in video tasks using an optical flow-based warping loss during training~\cite{lai2018, cao2021, li2021}. 
Used in isolation, however, this approach fails to generalize well to unseen data, often requiring fine-tuning for each new dataset. But it does have the advantage of not relying on camera poses or external structures. Lai~\etal~\cite{lai2018} and Cao~\etal~\cite{cao2021} pair a warping loss with a convolutional recurrent module which allows the network to learn temporal affinities more effectively. 

In fact, the use of recurrent structures is another way of enforcing temporal consistency in video tasks~\cite{kumar2018, zhang2019, patil2020}. Zhang~\etal~\cite{zhang2019} propagate spatio-temporal depth information across frames using a novel convolutional LSTM module which is trained using an adversarial loss. While this loss does not have any explicit temporal constraint like the warping loss, it is nonetheless shown to improve depth consistency. Patil~\etal's~\cite{patil2020} work on depth prediction and completion demonstrates that recurrent modules enforce consistency and improve accuracy even without special loss functions. Convolutional LSTMs, however, can be difficult to train due to their large memory requirements.

\textbf{Depth Fusion:} Fusion methods~\cite{newcombe2011, choe2021, weder2020, weder2021} achieve scene reconstruction by blending weighted signed distance field (SDF) volumes for each frame. The use of SDF volumes makes them scale poorly to large scenes and high resolutions. Keller~\etal~\cite{keller2013} and Lefloch~\etal~\cite{lefloch2015} propose point clouds as an alternative to SDFs. Traditional fusion methods, however, are not directly applicable to depth reconstruction as their focus is geometric reconstruction over multiple frames. Completeness, in the sense of reconstructing each visible point in a frame, is not a requirement. 

%% file: src/3.method.tex
\section{Consistent Video Depth Estimation}
\label{sec-method}
Given a sequence of RGB images $\obv{c}$ with known poses, our goal is to estimate a depth map $\obv{d}$ for each $\obv{c},~t\in\{0, 1, 2, ...\}$ such that the geometric representation of all scene points visible at time $t$ is accurate and consistent from $d^{t-1}$ to $\obv{d}$. 
Furthermore, we want to solve this problem online; that is, at any time instant $t$ future frames $d^i,~i>t$ are not known. Accuracy subsumes consistency since a 100$\%$ accurate reconstruction is also 100$\%$ consistent, but not vice versa. As a result, the majority of past depth estimation methods have focused on optimizing accuracy of reconstruction~\cite{cheng2020, tankovich2021, lipson2021, xu2022}. In practice, however, a perfectly accurate reconstruction is hardly ever achievable such that most of these methods suffer from inconsistency.

A simple way of enforcing temporal consistency is to use the known camera pose to elevate a depth map $\obv{d}$ to a 3D point cloud and then reproject it into future frames. However, such a reprojection-based approach suffers from the following shortcomings: it fails to handle dynamic scenes, it creates holes in disoccluded regions, and it will propagate any errors in $\obv{d}$ to all subsequent frames.

Our proposed approach seeks to exploit the advantages of a global point cloud while addressing the above-mentioned shortcomings. We do this in three steps: 
\begin{enumerate*}[label=(\textbf{\roman*})]
\item The temporal fusion step (Sec.~\ref{sec:temporal-filtering}) identifies and updates regions of the scene that changed from frame $t-1$ to $t$ by blending $\obv{d}$ with prior depth  from the global point cloud in image space.
\item Spatial fusion (Sec.~\ref{sec:depth-fusion}) recovers spatial details and corrects any errors in the blended output depth map from the first stage.
\item Finally, we update the global point cloud to incorporate changes from frame $t$ (Sec.~\ref{sec:global-pc-update}). 
\end{enumerate*}

%% file: src/3a.temporal-filtering.tex
\subsection{Temporal Fusion}
\label{sec:temporal-filtering}

\input{figures/temporal-filtering}

Let $\mathcal{P}\in\mathbb{R}^3$ be a prior 3D point cloud representation of a scene at time $t$ generated using the depth input from previous frames. For each $\vect{x}\in \mathcal{P}$ let $\varsigma(\vect{x})\in\mathbb{R}^3$ represent the RGB color, and $\rho(\vect{x})\in\mathbb{R}$ represent a confidence estimate of the point. Given the camera pose $\mat{T}_t$ and intrinsics matrix $\mat{K}$,  we define $\prior{d}$, $\prior{c}$ and $\prior{w}$ as the $H\times W$ depth, color and confidence projection, respectively, of $\mathcal{P}$.

The goal of the temporal fusion stage is to generate a 2D mask $\alpha\in[0, 1]^{H\times W}$ for the dynamic regions in $\prior{d}$. Specifically, we want $\alpha$  to identify visible points that change position from frame $t-1$ to $t$ solely due to motion, and not because of camera movement or uncertainty in the depth estimation. We then compute a fused depth map $\dtemp$ with the dynamic parts updated from $\obv{d}$ while preserving consistent depths from prior frames in the static regions:
\begin{align}
    \dtemp = \alpha\, \obv{d} + (1 -\alpha)\,\prior{d}
    \label{eqn:alpha-blending-one}
\end{align}
The blending mask $\alpha$ may be computed as the residual between consecutive frames warped using rigid flow. We observe, however, that a hand-crafted threshold for the residual does not generalize to all scenes, and is prone to failure in the presence of noise and depth artifacts. Thus, we use $\alpha=\Theta(\obv{d}, \prior{d}, \obv{c}, \prior{c})$ instead, where $\Theta(\cdot)$ is a convolutional neural network~(Fig.~\ref{fig:temporal-filtering}).  
We want $\alpha$ to have a strong bias towards the prior depth map in static regions. This is encouraged by using the ground truth --- rather than the generated --- prior point cloud $\mathcal{P}$ to render $\prior{c}$ and $\prior{d}$ during training. Using an L1 loss on the blended output $\dtemp$ then ensures that $\alpha=1$ minimizes the loss in scene regions that do not change from $t-1$ to $t$. As observed earlier, training $\Theta(\cdot)$ to directly output $\dtemp$ does not guarantee consistency due to CNN aliasing and rotational invariance. 

For the very first frame $t=0$, we use $\prior{d} = d^t$. Subsequently, we render the point cloud $\mathcal{P}$ by splatting each $\vect{x}\in\mathcal{P}$, along with color $\varsigma(\vect{x})$ and confidence $\rho(\vect{x})$, to the nearest rounded pixel with a Z-buffer check. While this introduces aliasing artifacts, it is much faster than surfel-based methods~\cite{pfister2000}. We ameliorate the aliasing through supersampling, and we use Rosenthal and Linsens's method~\cite{rosenthal2008} to fill holes and remove background points.

%% file: figures/temporal-filtering.tex
\begin{figure}[t]
  \centering
   \includegraphics[width=0.95\linewidth, trim={8.55cm, 7.2cm, 11.5cm, 6.25cm}, clip]{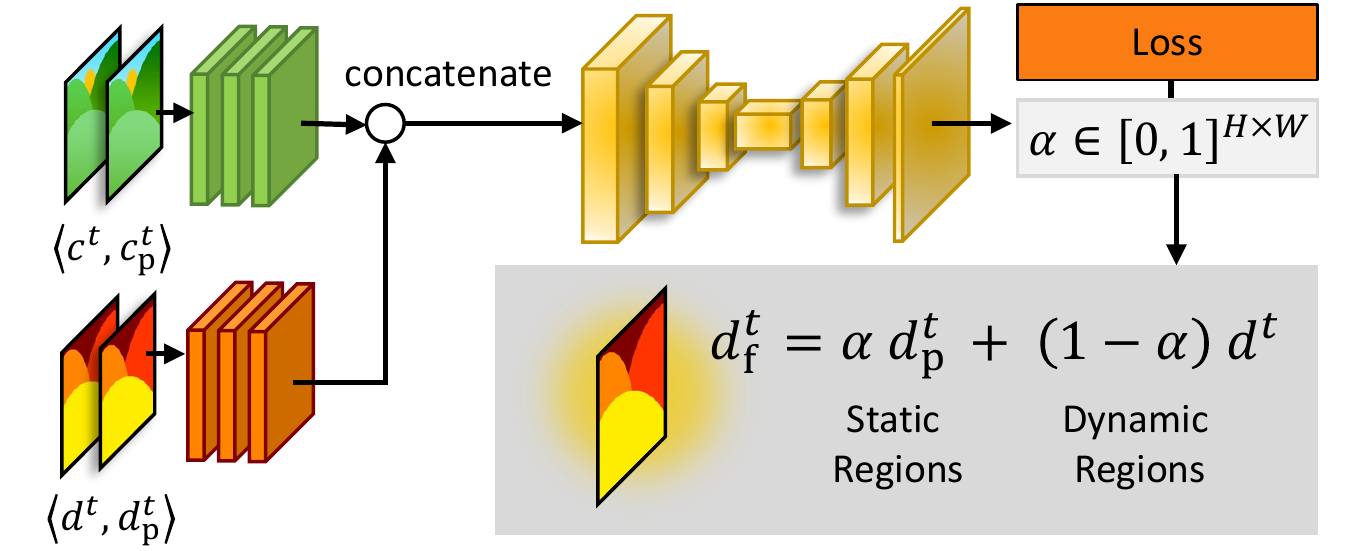}
   \caption{The temporal fusion stage generates a 2D mask $\alpha$ for scene regions that change due to motion in frame $t$. This mask is used to update the reprojected point cloud depth $\prior{d}$ with dynamic changes from the observed depth $d^t$.}
   \label{fig:temporal-filtering}
   \vspace{-4mm}
\end{figure}

%% file: src/3b.depth-fusion.tex
\subsection{Spatial Fusion}
\label{sec:depth-fusion}

\input{figures/depth-fusion}
\input{figures/network-outputs}

Using the ground truth to generate $\prior{d}$ allows for easily training the temporal fusion stage to maintain consistency in static regions. However, the ground truth is not available at inference-time so that the prior point cloud $\mathcal{P}$ --- and consequently $\prior{d}$ --- is likely to have errors from imperfect depth estimation in previous frames. We propose to address this problem in the spatial fusion stage.

The goal here is to assign a pixel-wise weight to $\obv{d}$ and $\dtemp$ based on a confidence estimate of the accuracy of each depth. This is then used to fuse the two depths in image space to generate the final output $\dout$:
\begin{align}
    \dout = \frac{\beta\,\dtemp + \gamma\, \obv{d}}{\beta + \gamma}
    \label{eqn:fusion}
\end{align}
We estimate the confidence weights $\beta, \gamma\in\mathbb{R}^{H\times W}$ using a neural network $\Phi(\cdot)$ (Fig.~\ref{fig:depth-fusion}). Note that while formulated similar to Eq.~\ref{eqn:alpha-blending-one}, the weights in Eq.~\ref{eqn:fusion} serve a different purpose and, hence, require a different training procedure for $\Phi(\cdot)$. In particular, we represent the weights as the input-dependent aleatoric uncertainty~\cite{kendall2017} which can be learnt in a self-supervised manner. The confidence $\gamma$ for $\obv{d}$ is then inversely related to the uncertainty as,
\begin{align}
    \gamma = \mathrm{exp}\big[-\Phi(d^t, c^t)]
    \label{eqn:gamma}
\end{align}
The confidence $\beta$ for $\dtemp$, in addition to the aleatoric uncertainty $\Phi(\dtemp)$, also incorporates the prior confidence $\prior{w}$ weighted by the temporal mask $\alpha$:
\begin{align}
    \beta = \alpha\,\prior{w} \mathrm{exp}\big[-\Phi( \dtemp, \obv{c})\big]
    \label{eqn:beta}
\end{align}

We box-filter $\prior{w}$ before computing $\beta$ to suppress any high-frequency artifacts from the splatting process. Note that $\Phi(\cdot)$ can model uncertainty in $\dtemp$ resulting from errors in temporal filtering, so that Eq.~\ref{eqn:fusion} is not the same as $\alpha$-blending $d^t$ with $\prior{w}\,\mathrm{exp}\big[-\Phi(\prior{d}, \obv{c}) \big]\,\prior{d}$.

%% file: figures/depth-fusion.tex
\begin{figure}[t]
  \centering
   \includegraphics[width=0.95\linewidth, trim={8.45cm, 6.5cm, 11.75cm, 6.0cm}, clip]{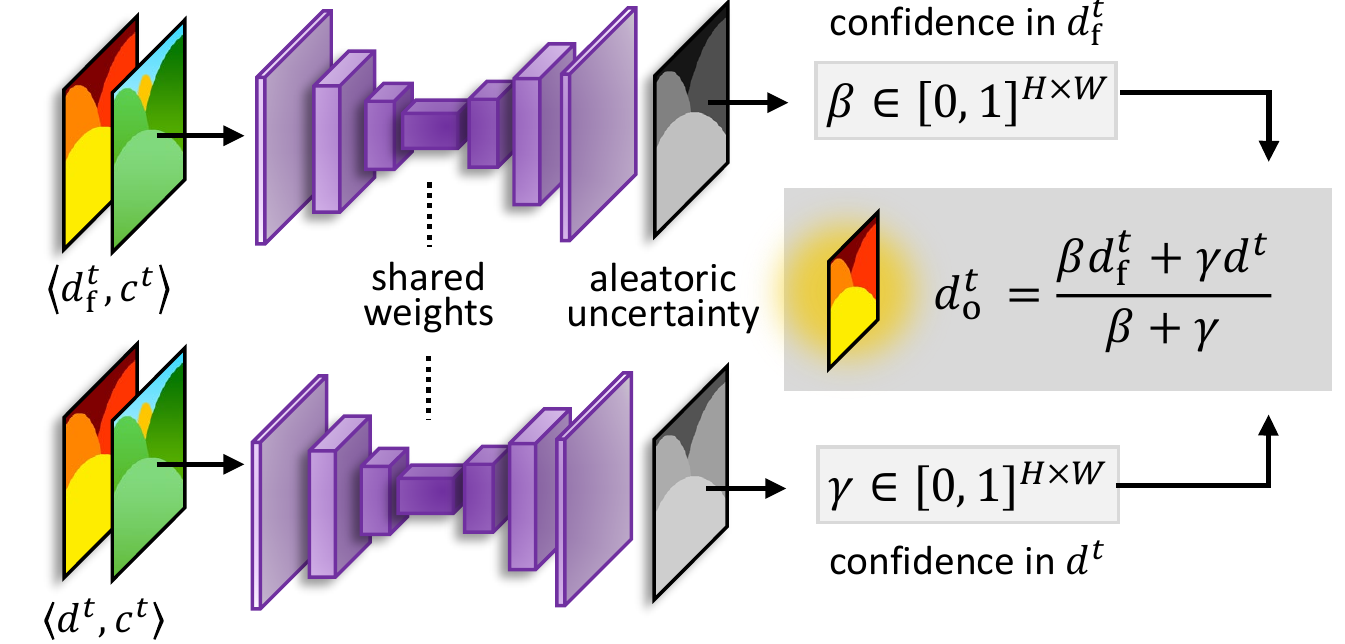}
   \caption{The spatial fusion stage uses the observed depth $\obv{d}$ to improve static regions in the temporally fused  depth $\dtemp$ based on confidence in the accuracy of both depths. The confidence is estimated as the learnt uncertainty in the depth prediction model. }
   \label{fig:depth-fusion}
\end{figure}

%% file: figures/network-outputs.tex
\begin{figure}[t]
  \centering
   \includegraphics[width=1.0\linewidth, trim={0.0cm, 13.0cm, 0.0cm, 1.5cm}, clip]{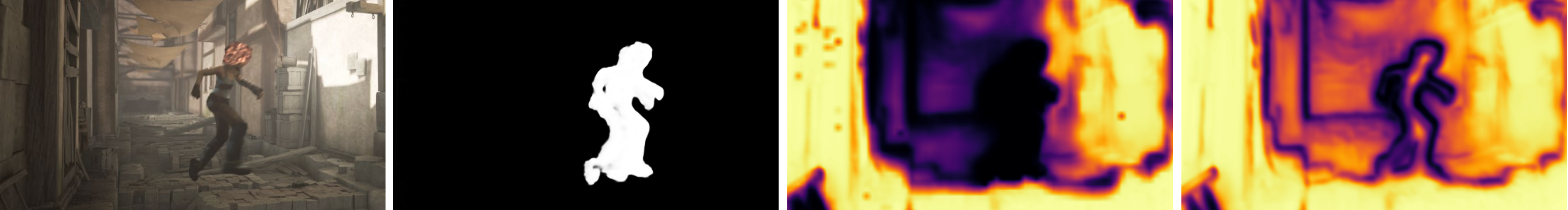}
   \caption{\textit{Left to right}: The input $c^t$, the dynamic change mask $\alpha$ (Eq.~\ref{eqn:alpha-blending-one}), the confidence estimate $\beta$ (Eq.~\ref{eqn:beta}), and the confidence estimate $\gamma$ (Eq.~\ref{eqn:gamma}). \footnotesize{Original image \textcopyright~ Blender Foundation~$|$~www.sintel.org}}
   \label{fig:network-outputs}
   \vspace{-4mm}
\end{figure}

%% file: src/3c.global-pc-update.tex
\input{tables/results-mono-improvement-with-frames}

\subsection{Global Point Cloud Update}
\label{sec:global-pc-update}

In a final step, we update the global point cloud $\mathcal{P}\subseteq\mathbb{R}^3$ to reflect changes in the scene from frame $t$. This includes updating the confidence $\rho(\vect{x})$, and the color $\varsigma(\vect{x})$ of all points $\vect{x}\in\mathcal{P}$ observed at $t$. This is similar to the TSDF volume integration step in traditional fusion methods~\cite{izadi2011, newcombe2011}. However, here we have a discrete point cloud representation of the scene which cannot be directly averaged like a TSDF. Therefore, achieving a similar effect involves: \begin{enumerate*}[label=(\roman*)]
\item associating points in $\mathcal{P}$ with pixels in $\dout$, and updating the position, confidence and color of points for which a correspondence is found;
\item adding pixels from $\obv{d}$ for which no association is found as newly observed points into $\mathcal{P}$; and
\item removing points with low confidence from $\mathcal{P}$.
\end{enumerate*}

Let $\obv{z} \in\mathbb{R}^{H\times W \times 3}$ represent the inverse projection of depth map $\obv{d}$ to 3D. Updates to each $\vect{x}\in\mathcal{P}$ are computed in continuous image space to minimize quantization artifacts from using a point cloud. Thus, we update a given $\vect{x}$ by sampling
$\obv{z}$ and the color image $\obv{c}$ at the projected 2D location $\hat{\vect{x}}$ of $\vect{x}$ on the image plane. 
We denote the sampled values as $\obv{c}[\hat{\vect{x}}]\in \mathbb{R}^3$ and $\obv{z}[\hat{\vect{x}}]\in \mathbb{R}^3$, respectively, using square brackets to represent bilinear sampling of the image at the indexed location. We only update unoccluded points identified by $\alpha[\hat{\vect{x}}] < 0.5$, where $\alpha$ is from Eq.~\ref{eqn:alpha-blending-one}. The updated $\vect{x}$, color $\varsigma(\vect{x})$ and confidence $\rho(\vect{x})$ are computed as,
\begin{align}
    \vect{x} &= \frac{\beta[\hat{\vect{x}}] \cdot \vect{x} + \gamma[\hat{\vect{x}}]\cdot \obv{z}[\hat{\vect{x}}]}{\beta[\hat{\vect{x}}] + \gamma[\hat{\vect{x}}]} \\
    \varsigma(\vect{x}) &= \frac{\beta[\hat{\vect{x}}]\cdot \varsigma(\vect{x}) + \gamma[\hat{\vect{x}}]\cdot \obv{c}[\hat{\vect{x}}]}{\beta[\hat{\vect{x}}] + \gamma[\hat{\vect{x}}]} \\
    \rho(\vect{x}) &= \beta[\hat{\vect{x}}] + \gamma[\hat{\vect{x}}]
\end{align}
where $\beta,~\gamma \in \mathbb{R}^{H \times W}$ are the fusion weights from Eqs.~\ref{eqn:gamma} and~\ref{eqn:beta}. For any $\vect{x}$ outside the field-of-view, or which is occluded, we set $\rho(\vect{x}) = \rho(\vect{x}) - 1$. We identify newly observed points as pixel locations in $\obv{z}$ for which $\alpha \geq 0.5$, and add them to $\mathcal{P}$. Finally, we remove all points $\vect{x}\in\mathcal{P}$ for which $\rho(\vect{x}) < \epsilon$. We find  $\epsilon=3\times10^{-2}$ to work well empirically.

%% file: tables/results-mono-improvement-with-frames.tex
\setlength{\tabcolsep}{5pt}

\begin{table*}
\begin{center}
\begin{tabular}{lccccccccccccc}
\toprule
\multicolumn{14}{c}{MPI Sintel Dataset} \\
\midrule
 & OPW$\downarrow$ & SC$\downarrow$ & RTC$\uparrow$ & TCM$\uparrow$ & TCC$\uparrow$ & SD(L1)$\downarrow$ & & RAE$\downarrow$ & RMS$\downarrow$ & $\delta_1\uparrow$ & $\delta_2\uparrow$ & $\delta_3\uparrow$ & $N$\\

\midrule
\small{MiDaS~\cite{ranftl2020}} & 0.611 & 0.675 & 0.211 & 0.233 & 0.413 & 0.597 & &	0.279 &	3.013 &	0.610 &	0.798 &	0.896 & 1 \\
\small{WSVD~\cite{wang2019}$^\ddagger$} & 0.613 & 0.704 & 0.288 & 0.356 & 0.449 & 0.654 & &	0.360 &	3.775 &	0.466 &	0.705 &	0.825 & 2 \\
\small{DPT~\cite{ranftl2021}} & 0.424 & 0.493 & 0.320 & 0.417 & 0.482 & 0.539 & & 0.224 &	2.678 &	0.686 &	0.864 &	0.932 & 1 \\
\midrule
\small{Ours-DPT$^{\dag \ddagger}$} & 
\makecell{\textbf{0.255}} & 
\makecell{\textbf{0.295}} & 
\makecell{\textbf{0.489}} & 
\makecell{\textbf{0.470}} & 
\makecell{\textbf{0.559}} & 
\makecell{\textbf{0.474}} & &
\makecell{\textbf{0.197}} & 
\makecell{\textbf{2.400}} & 
\makecell{\textbf{0.710}} & 
\makecell{\textbf{0.885}} & 
\makecell{\textbf{0.942}} & 
1 \\

\midrule
\multicolumn{14}{c}{ScanNet Dataset} \\
\midrule
\small{ST-CLSTM~\cite{zhang2019}$^\ddagger$} & 0.087 & 0.087 & 0.310 & 0.134 & 0.400 &	0.076 & &	0.289 &	0.751 & 0.772 &	0.932 &	0.983 & 5 \\
\small{TCMD~\cite{li2021}$^\ddagger$} & 0.031 & 0.031 & 0.628 & 0.166 & 0.548 & 0.083 & & 0.300 &	0.811 & 0.740 & 0.927 & 0.981 & 1 \\
\small{N-RGB$\rightarrow$D~\cite{liu2019}$^{\dag \ddagger}$} & 0.043 & 0.042 & 0.557 & 0.169 &	0.482 & 0.088 & & 0.285 & 0.674 & 0.814 & 0.952 & 0.983 & 1 \\
\small{Demon~\cite{ummenhofer2017}} & 0.160 & 0.159 & 0.143 & 0.107 & 0.271 & 0.098 & & 0.228 & 0.651 & 0.839 & 0.951 &	0.982 & 2 \\
\small{DeepV2D~\cite{teed2018}} & 0.038 & 0.038 & 0.483 & 0.143 & 0.495 & 0.058 &	 & \textbf{0.153} &	 \textbf{0.502} & 0.957 & 0.995 & \textbf{0.999} & 2 \\

\small{DPT~\cite{ranftl2021}} & 0.033 & 0.033 & 0.540 & 0.141 & 0.536 & 0.055 & &	0.213 &	0.582 &	0.971 & \textbf{0.996} & \textbf{0.999} & 1 \\
\midrule
\small{Ours-DPT$^{\dag \ddagger}$} &
\makecell{\textbf{0.011} } &
\makecell{\textbf{0.010} } &
\makecell{\textbf{0.863} } & 
\makecell{\textbf{0.179} } & 
\makecell{\textbf{0.639} } &
\makecell{\textbf{0.052} } & &
\makecell{{0.210} } &
\makecell{{0.576} } &
\makecell{\textbf{0.974} } &
\makecell{\textbf{0.996} } &
\makecell{\textbf{0.999} } & 
1
\\
\bottomrule
\noalign{\vskip 1mm} 
\multicolumn{14}{r}{$^\dag$\footnotesize{Method uses temporal data structure.} $^\ddagger$ \footnotesize{Method has temporal consistency constraints.}} \\
\end{tabular}
\vspace{-3mm}
\caption{Comparing our method using a DPT backbone to SOTA monocular depth methods. $N$ is the number of frames required at inference. Our method achieves the best results overall with significantly higher consistency, and similar or better accuracy. }
\label{table:results-mono}
\end{center}
\vspace{-4mm}
\end{table*}

%% file: src/3d.training.tex
\section{Training Procedure}
\label{sec:training}
We train the temporal fusion network $\Theta(\cdot)$ using samples $(\obv{d}, \obv{c}, \prior{d}, \prior{c})$ where $(\obv{c}, \obv{d})$ are the color image and estimated depth of the current frame $t$, and $(\prior{d}, \prior{c})$ are generated by reprojecting the ground truth depth and color from frame $t-k$ to $t$, for some arbitrary $k$ between -7 and 7. We use future frames ($k < 0$) only during training as a data augmentation technique --- during inference, our method only sees the current frame $t$ and the global point cloud.

$\Theta(\cdot)$ consists of two sets of three residual layers followed by a four-layer U-Net~\cite{ronneberger2015}. The residual layers extract features from the concatenated depths $\langle \prior{d}, \obv{d} \rangle$ and concatenated color images $\langle \prior{c}, \obv{c} \rangle$ which are then fed into the U-Net to generate the blending mask $\alpha$~(Fig.~\ref{fig:temporal-filtering}). This network architecture is much simpler and has fewer parameters than either Mask-RCNN~\cite{he2017} or optical flow methods, both of which have been previously used ~\cite{luo2020, yoon2020, li2021temp} to mask dynamic elements of a scene (Table~\ref{table:filtering-networks-comparison}). We use ReLU activations on all but the last layer of the U-Net, which has a sigmoid. As batch normalization adds a depth-dependent bias~\cite{weder2020}, we use layer normalization instead.

The motivation behind using the ground truth rather than the estimated depth for $\prior{d}$ is to encourage a strong bias towards the prior depth ($\alpha=0$) in static regions. This ensures $\Theta(\cdot)$ only identifies changes due to motion. The reprojected ground truth remains optimal in regions that do not change from frame $t-k$ to $t$. Thus, an L1 loss on $\dtemp$ encourages $\alpha=0$ in these parts. For dynamic objects, the prior depth may or may not be optimal based on the quality of the estimated depth, and the network learns a suitable blending weight. We found a binary cross entropy (BCE) loss on $\alpha$ improves convergence and pushes it closer to binary:
\begin{align}
\mathrm{L}_{\mathrm{BCE}} = \mathrm{BCE}\Big(\alpha, \mathds{1} \big(\, \lVert \obv{d} - \obv{g} \rVert_1 < \lVert \prior{d} - \obv{g} \rVert_1 \big)\Big)
\end{align}
where $\mathds{1}(\cdot)$ is the indicator function of the pixels satisfying the  inequality and $\obv{g}$ is the ground truth depth. We also include an L1 loss $\mathrm{L}_{\mathrm{G}}$ on the depth gradients, and a VGG loss on the fused depth. We found the latter helps capture fine detail better than an L1 loss alone. Thus, we supervise the temporal fusion network with loss: 
\begin{align}
    \mathrm{L}_{\Theta}= \lVert \dtemp - \obv{g} \rVert_1 + \mathrm{L}_{\mathrm{BCE}} + \mathrm{L}_{\mathrm{G}} + \mathrm{VGG}(\dtemp, \obv{g}) 
\end{align}
We weigh the terms as 10, 0.1, 0.05 and 0.05, respectively.
We implement the depth fusion network $\Phi(\cdot)$ as a four-layer U-Net with instance normalization, and ReLU activations on all layers. We train the network using a self-supervised loss on the uncertainty $\obv{s} = \Phi(\obv{d}, \obv{c})$ of the input depth:
\vspace{-2mm}
\begin{align}
    \mathrm{L}_{\Phi} = \frac{1}{m} \left\Vert\, \mathrm{exp}(-\obv{s})\,|\,\obv{d} - \obv{g}| + \lambda_s \obv{s}\, \right\Vert_1
\end{align}
where $m$ is the number of pixels and the second term is a regularizer to prevent the trivial $\obv{s} = \infty$. We use $\lambda_s=0.03$. As Kendall~\etal~\cite{kendall2017} show, this loss term minimizes the log variance of a Gaussian likelihood model for the uncertainty. 

We train both networks on the FlyingThings3D dataset~\cite{mayer2016}. This consists of rendered sequences of objects moving along random 3D trajectories, and has ground truth depth and camera poses. For fine-tuning we render 50 60-frame sequences from a custom Blender scene with human-like motion in an indoor setting. We estimate $\obv{d}$ with RAFT-Stereo~\cite{lipson2021} and use the inverse depth for training.

%% file: src/4.experiments.tex
\begin{figure}[t]
  \centering
   \includegraphics[width=1.0\linewidth, trim={0cm, 9.5cm, 0cm, 3cm}, clip]{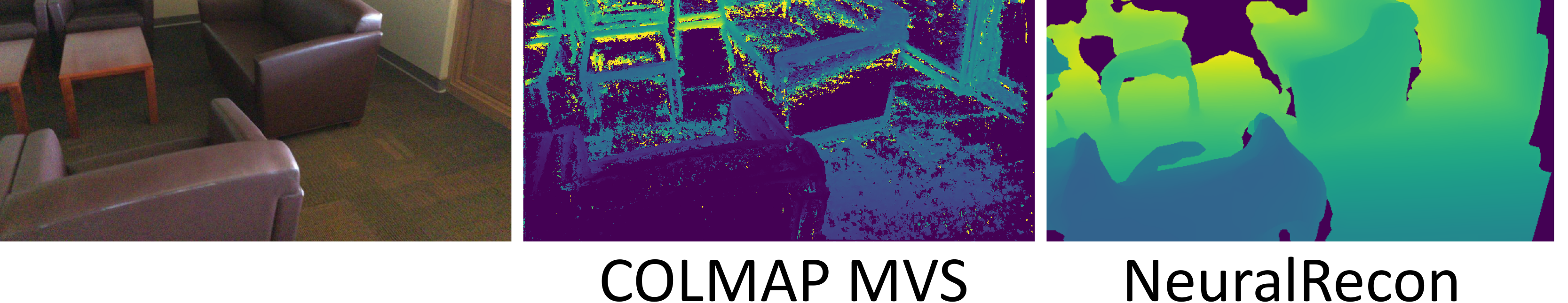}
   \caption{MVS and fusion methods like NeuralRecon~\cite{sun2021} suffer from holes in the output making them ill-suited to AR applications. }
   \label{fig:mvs-tsdf-holes}
   \vspace{-3mm}
\end{figure}

\section{Experiments}
\label{sec:experiments}

\subsection{Implementation Details}
Both networks are implemented in PyTorch and trained on six Nvidia Tesla V100 GPUs. The temporal fusion network is trained for five epochs on 500$\times$500 crops from the FlyingThings3D dataset in batches of 18. We use a learning rate of $5\times10^{-4}$ and a weight decay of 1$\times 10^{-4}$. The spatial fusion network is trained on the depth output of the temporal filtering stage $\dtemp$ using the FlyingThings3D dataset with a similar batch and input size, and a learning rate of 1$\times 10^{-4}$. We fine-tune both networks on 320$\times$320 crops of our Blender data for 14 epochs at a learning rate of 1$\times 10^{-6}$.

\input{tables/results-cvd}

\subsection{Testing Datasets}
\noindent
\textbf{MPI Sintel}: The MPI Sintel dataset~\cite{butler2012, wulff2012} consists of long synthetic stereo sequences with large motion and depth range. We evaluate both stereo and monocular depth estimation on the final pass of the 22 ``training scenes'' which have ground truth optical flow, pose, and depth. Given the extremely large motion and depth of some scenes, we follow Teed~\etal~\cite{teed2021} and restrict our evaluation to pixels with flow $\leq$ 250 pixels and depth $\leq$ 30 meters.

\noindent
\textbf{ScanNet}: The ScanNet dataset~\cite{dai2017} consists of video sequences of static indoor scenes captured with an RGB-D sensor and annotated with camera poses. We evaluate the monocular variant of our method on 20 60-frame sequences from different scenes. ScanNet does not have ground truth optical flow, so we estimate it using RAFT~\cite{teed2020}.

\input{tables/results-stereo-improvement}
\input{figures/results-epi-stereo}

\subsection{Baselines}
Our choice of baselines is motivated by the use cases we envision for an online method. These include low-latency AR/VR applications~\cite{chaurasia2020, xiao2022} requiring temporal consistency, completeness~(\ie no holes), and the handling of dynamic objects. Hence, we evaluate stereo depth methods~\cite{lipson2021, tankovich2021} on the MPI Sintel dataset, and monocular methods~\cite{li2021, liu2019, ranftl2021, ranftl2020, wang2019, zhang2019} on both the MPI Sintel and ScanNet datasets. Additionally, we evaluate learned structure-from-motion methods~\cite{ummenhofer2017, teed2018}. We do not compare with traditional multi-view stereo or TSDF fusion methods as they suffer incompleteness, and fail in dynamic scenes~(Fig.~\ref{fig:mvs-tsdf-holes}). 

For each baseline, we use pre-trained models provided by the authors and evaluate it only on the dataset it has originally been tested on --- with the only exception being models trained on the NYU-V2 dataset~\cite{silberman2012}. We evaluate these models on ScanNet instead (NYU-V2 is a similar sensor-captured RGB-D dataset, but does not have ground truth poses). We use the efficient \textit{Hybrid} variant of Ranftl~\etal's DPT~\cite{ranftl2021} in all experiments, and the five-layer \textit{3D-GAN} version of Zhang~\etal's ST-CLSTM~\cite{zhang2019} .

\subsection{Evaluation Metrics}
Given a sequence of estimated depth maps $d^0, d^1, ..., d^n$ along with color images $c^0, c^1, ..., c^n$ and ground truth $g^0, g^1, ..., g^n$, we evaluate depth estimation quality both temporally and spatially. A common metric for temporal consistency of video uses the optical flow $O_{t\rightarrow t+1}$ to evaluate the change in a pixel's depth from frame $t$ to $t+1$~\cite{lai2018, cao2021, wang2022}. Let $d_w^{\,t+1}$ be the estimated depth $d^{\,t+1}$ backward warped into frame $t$ using $O_{t\rightarrow t+1}$. Then the optical flow-based warping error (OPW) is defined as:
\begin{align}
\textrm{OPW} = \frac{1}{n-1} \sum_t \lVert M^t\, (d_w^{\,t+1} - d^t) \rVert_1
\end{align}
where $M^t=\mathrm{exp}(-\kappa(c_w^{\,t+1} - c^t))$ is a pixel-wise occlusion mask. Intuitively, OPW is the average meter change in the depth of a point across frames. It is expected to be higher for the outdoor scenes in MPI-Sintel than the indoor ScanNet dataset. Li~\etal~\cite{li2021} present a variant of this error known as the relative temporal consistency (RTC) metric, which is shown to align with humans' perception of consistency: 
\begin{align}
\textrm{RTC} = \frac{1}{mn} \sum_t \left\Vert \mathds{1} \Big( M^t\,\textrm{max} \big(\frac{d_w^{t+1}}{d_t}, \frac{d_t}{d_w^{t+1}} \big) < \tau \Big) \right\Vert_0
\end{align}
where $m$ is the number of pixels, $\mathds{1}(\cdot)$ is the indicator function of the set of pixels satisfying the enclosed inequality, and the $0$-norm measures the number of non-zero pixel values. We set $\kappa=50$ and $\tau=1.01$ for our evaluation.

A disadvantage of OPW and RTC is that both metrics are optimized by the degenerate case $d^t = k~\forall~t\in\{0, 1, ..., n\}$ for any constant $k$. Hence we augment OPW and RTC with the temporal change consistency (TCC) and the temporal motion consistency(TCM) metrics of Zhang~\etal~\cite{zhang2019}:
\begin{align}
\textrm{TCC} &= \frac{1}{n-1} \sum_t \textrm{SSIM}\big(\,\lvert d^t - d^{t+1} \rvert, \lvert g^t - g^{t + 1}\rvert\,\big) \\
\textrm{TCM} & = \frac{1}{n-1} \sum_t \textrm{SSIM}(O(d^t, d^{t+1}), O(g^t, g^{t + 1})) 
\end{align}
where SSIM is the structural self-similarity~\cite{wang2004}, and $O$ is the optical flow $O_{t\rightarrow t+1}$. In addition, we propose the self-consistency (SC) metric which uses the estimated depth $d^t$ to compute $d_w^{t+1}$ instead of the optical flow. Like OPW, it measures the average meter change in depth frame-to-frame. Finally, as in Long~\etal~\cite{long2021} we measure the standard deviation of the L1 error SD(L1) across all frames. 

For the spatial error we use the standard metrics: relative absolute error (RAE); root mean squared error (RMS); and the percentage of bad pixels $\delta_1$, $\delta_2$, $\delta_3$ at thresholds $1.25$, $1.25^2$ and $1.25^3$, respectively~\cite{zhang2019}. We align the affine-invariant monocular methods for evaluation by using the ground truth to calculate a scale and shift factor~\cite{ranftl2020}.

\subsection{Results}
\input{figures/results-mono}

\input{tables/ablation-pipeline}

\input{tables/temporal-networks-compare}

Table~\ref{table:results-mono} presents the quantitative evaluation of our method and the monocular baselines, with a qualitative comparison in Fig.~\ref{fig:results-qualitative-mono}. 
Across both datasets, our depth has a 50$\%$ lower OPW error than the next best method. Intuitively, this means for an average scene point, the frame-to-frame meter difference in depth is less than half that of the other method. Furthermore, the structural similarity of the temporal change in our depths and ground truth (TCC) is 15$\%$ higher, indicating that dynamic regions are not simply averaged across frames. A similar trend is seen in the improved performance of the stereo methods in Table~\ref{table:results-stereo} with an improvement in OPW and TCC of around $30\%$ and $5\%$ respectively. Fig.~\ref{fig:results-qualitative-stereo} provides a qualitative visualization of the stereo results. The improvement from our method is compared to Luo~\etal's CVD ~\cite{luo2020} in Table~\ref{table:results-cvd}. CVD runs offline, fine-tuning the backbone network for $\sim$20 minutes at inference-time. As such, it represents the high bar for temporal consistency. Our method achieves a good portion of the gains of CVD while being generalizable and online. 

Table~\ref{table:filtering-networks-comparison} lists the size and runtime of our two networks, comparing $\Theta(\cdot)$ to RAFT~\cite{teed2020} and Mask-RCNN~\cite{he2017} --- the latter having been used for dynamics filtering~\cite{li2021, yoon2020, luo2020}. 
Table~\ref{table:ablation} show how ablating each stage affects the performance of our method. For \textit{No Temporal Fusion} we set $\alpha=0$ in Eq.~\ref{eqn:alpha-blending-one}; \textit{No Spatial Fusion} uses $\dout = \dtemp$; and \textit{No Global PC} does not use a global point cloud, using the reprojected points from frame $t-1$ only. 
The temporal fusion stage has the strongest effect on consistency and accuracy. Setting $\alpha=0$ effectively averages the frames via Eq.~\ref{eqn:fusion} yielding temporally smooth --- albeit erroneous --- output. Hence, the drop in TCC but the lower OPW error as the latter is minimized when depth variance across frames is low.

%% file: tables/results-cvd.tex
\setlength{\tabcolsep}{5pt}

\begin{table*}
\begin{center}
\begin{tabular}{lccccccccccccc}
\toprule
\multirow{2}{*}{} & \multicolumn{12}{c}{MPI Sintel Dataset} \\
\midrule
 & OPW$\downarrow$ & SC$\downarrow$ & RTC$\uparrow$ & TCM$\uparrow$ & TCC$\uparrow$ &  SD(L1)$\downarrow$ & & RAE$\downarrow$ & RMS$\downarrow$ & $\delta_1\uparrow$ & $\delta_2\uparrow$ & $\delta_3\uparrow$ & $N$\\

\midrule
\small{MC~\cite{li2019}} & 0.553 &	0.608 & 0.233 & 0.423 & 0.402 & 	0.619 &	& 0.371 &	3.771 &	0.462 &	0.690 &	0.820  & 1 \\
\midrule
\small{CVD-MC~\cite{luo2020}} &
\makecell{{0.226} \\[-0.1cm] \footnotesize{\color{mygreen}{-59.1$\%$}}} &	
\makecell{{0.245} \\[-0.1cm] \footnotesize{\color{mygreen}{-59.7$\%$}}} &
\makecell{{0.402} \\[-0.1cm] \footnotesize{\color{mygreen}{+72.5$\%$}}} & 
\makecell{{0.515} \\[-0.1cm] \footnotesize{\color{mygreen}{+21.7$\%$}}} & 
\makecell{{0.460} \\[-0.1cm] \footnotesize{\color{mygreen}{+14.4$\%$}}} &
\makecell{0.589 \\[-0.1cm] \footnotesize{\color{mygreen}{-4.85$\%$}}} &	& 
\makecell{{0.338} \\[-0.1cm] \footnotesize{\color{mygreen}{-8.89$\%$}}} &	
\makecell{{3.078} \\[-0.1cm] \footnotesize{\color{mygreen}{-18.4$\%$}}} &	
\makecell{{0.493} \\[-0.1cm] \footnotesize{\color{mygreen}{+6.71$\%$}}}& 
\makecell{{0.702} \\[-0.1cm] \footnotesize{\color{mygreen}{+1.74$\%$}}} & 
\makecell{0.819 \\[-0.1cm] \footnotesize{\color{red}{-0.12$\%$}}} &
All
\\

\small{Ours-MC} &
\makecell{0.377 \\[-0.1cm] \footnotesize{\color{mygreen}{-31.8$\%$}}} &	
\makecell{0.399 \\[-0.1cm] \footnotesize{\color{mygreen}{-34.4$\%$}}} &	
\makecell{0.360 \\[-0.1cm] \footnotesize{\color{mygreen}{+54.5$\%$}}} & 
\makecell{0.443 \\[-0.1cm] \footnotesize{\color{mygreen}{+4.73$\%$}}} & 
\makecell{0.423 \\[-0.1cm] \footnotesize{\color{mygreen}{+5.22$\%$}}} &
\makecell{{0.583} \\[-0.1cm] \footnotesize{\color{mygreen}{-5.82$\%$}}} &	& 
\makecell{0.361 \\[-0.1cm] \footnotesize{\color{mygreen}{-2.70$\%$}}} &	
\makecell{3.711 \\[-0.1cm] \footnotesize{\color{mygreen}{-1.59$\%$}}} &	
\makecell{0.466 \\[-0.1cm] \footnotesize{\color{mygreen}{+0.87$\%$}}}& 
\makecell{0.696 \\[-0.1cm] \footnotesize{\color{mygreen}{+0.87$\%$}}} & 
\makecell{{0.823} \\[-0.1cm] \footnotesize{\color{mygreen}{+0.37$\%$}}} &
1 \\

\bottomrule
\end{tabular}
\caption{We compare our approach to CVD using the Mannequin Challenge~(MC) backbone for both methods. The percentage change in each metric over the baseline (MC) is listed in color. Our method realizes a large portion of the gains of CVD, even though CVD fine-tunes the backbone network for each input ($\sim$20 mins) and requires all frames beforehand. Our method is trained once and runs online.}
\label{table:results-cvd}
\vspace{-4mm}
\end{center}
\end{table*}

%% file: tables/results-stereo-improvement.tex
\setlength{\tabcolsep}{5pt}

\begin{table*}
\begin{center}
\begin{tabular}{lcccccccccccc}
\toprule
\multicolumn{13}{c}{MPI Sintel Dataset} \\
\midrule
 & OPW$\downarrow$ & SC$\downarrow$ & RTC$\uparrow$ & TCM$\uparrow$ & TCC$\uparrow$ &  SD(L1)$\downarrow$ & & RAE$\downarrow$ & RMS$\downarrow$ & $\delta_1\uparrow$ & $\delta_2\uparrow$ & $\delta_3\uparrow$\\
 
\midrule
\small{RAFT-Stereo~\cite{lipson2021}} & 0.409 & 0.507 & 0.615 & 0.424 & 0.642 & 0.795 & & 0.239 & 31.26  & \textbf{0.915} & \textbf{0.944} & \textbf{0.961}\\

\small{Ours-RAFT-S.} & 
\makecell{0.268 \\[-0.1cm] \footnotesize{\color{mygreen}{-34.5$\%$}}} & 
\makecell{0.340 \\[-0.1cm] \footnotesize{\color{mygreen}{-32.9$\%$}}} &	
\makecell{\textbf{0.701} \\[-0.1cm] \footnotesize{\color{mygreen}{+14.0$\%$}}} &	
\makecell{0.492 \\[-0.1cm] \footnotesize{\color{mygreen}{+16.0$\%$}}} & 
\makecell{0.674 \\[-0.1cm] \footnotesize{\color{mygreen}{+4.98$\%$}}} & 
\makecell{0.745 \\[-0.1cm] \footnotesize{\color{mygreen}{-6.29$\%$}}} &	& 
\makecell{0.227 \\[-0.1cm] \footnotesize{\color{mygreen}{-5.02$\%$}}} & 
\makecell{21.86 \\[-0.1cm] \footnotesize{\color{mygreen}{-30.1$\%$}}} &	\makecell{\textbf{0.915} \\[-0.1cm] \footnotesize{\color{gray}{+0.00$\%$}}} & 
\makecell{0.943 \\[-0.1cm] \footnotesize{\color{red}{-0.11$\%$}}} & 
\makecell{0.959 \\[-0.1cm] \footnotesize{\color{red}{-0.21$\%$}}} \\

\midrule
\small{HITNet~\cite{tankovich2021}} & 0.225 & 0.264 & 0.599 & 0.441 & 0.641 & 0.341 & & 0.087 & 12.32 & 0.878 & 0.915 &	0.941 \\

\small{Ours-HITNet} &  
\makecell{\textbf{0.158} \\[-0.1cm] \footnotesize{\color{mygreen}{-29.8$\%$}}} & 
\makecell{\textbf{0.197} \\[-0.1cm] \footnotesize{\color{mygreen}{-25.4$\%$}}} & 
\makecell{0.675 \\[-0.1cm] \footnotesize{\color{mygreen}{+12.7$\%$}}} & 
\makecell{\textbf{0.506} \\[-0.1cm] \footnotesize{\color{mygreen}{+14.7$\%$}}} & 
\makecell{\textbf{0.675} \\[-0.1cm] \footnotesize{\color{mygreen}{+5.30$\%$}}} & 
\makecell{\textbf{0.316} \\[-0.1cm] \footnotesize{\color{mygreen}{-7.33$\%$}}} & & 
\makecell{\textbf{0.086} \\[-0.1cm] \footnotesize{\color{mygreen}{-1.15$\%$}}} & 
\makecell{\textbf{8.900} \\[-0.1cm] \footnotesize{\color{mygreen}{-27.8$\%$}}}  &	
\makecell{0.878 \\[-0.1cm] \footnotesize{\color{gray}{+0.00$\%$}}} & 
\makecell{0.916 \\[-0.1cm] \footnotesize{\color{mygreen}{+0.11$\%$}}} &	
\makecell{0.943 \\[-0.1cm] \footnotesize{\color{mygreen}{+0.21$\%$}}} \\

\bottomrule
\end{tabular}
\caption{Evaluating two SOTA stereo methods when used with our point cloud-based fusion approach. The percentage change is listed in color below our result for each metric. Our method leads to significant improvement in temporal consistency and preserves spatial quality.}
\label{table:results-stereo}
\end{center}
\vspace{-6mm}
\end{table*}

%% file: figures/results-epi-stereo.tex
\begin{figure*}[h]
  \centering
   \includegraphics[width=0.99\linewidth, trim={0.3cm, 4.80cm, 0.0cm, 0.4cm}, clip]{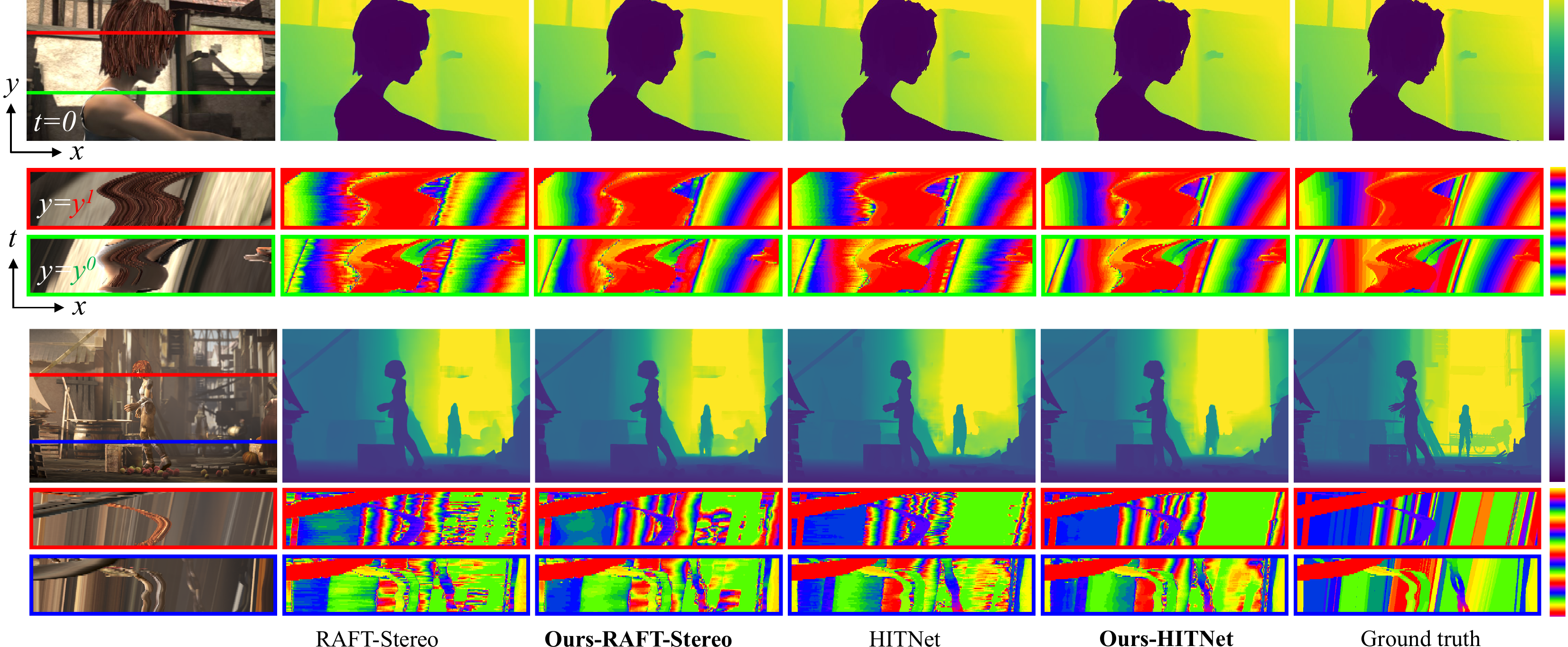}
   \caption{The improvement in consistency from our method, visualized as the variation in an epipolar plane image (EPI). An EPI represents a slice of the video in the $x$-$t$ dimensions for a fixed $y$ (indicated in \textcolor{mygreen}{green}, \textcolor{red}{red} and \textcolor{blue}{blue}). We use a high-gradient color-map to better visualize changes in the EPI. Our method  reduces high-frequency oscillation along $t$ indicating better frame-to-frame consistency, and preserves the spatial ($x$-$y$) quality of depth. We omit the axis labels in subsequent figures. \footnotesize{Original images \textcopyright~ Blender Foundation~$|$~www.sintel.org}}
   \label{fig:results-qualitative-stereo}
   \vspace{-4mm}
\end{figure*}

%% file: figures/results-mono.tex
\begin{figure*}[t!]
  \centering
   \includegraphics[width=0.99\linewidth, trim={10cm, 0cm, 0.0cm, 0.3cm}, clip]{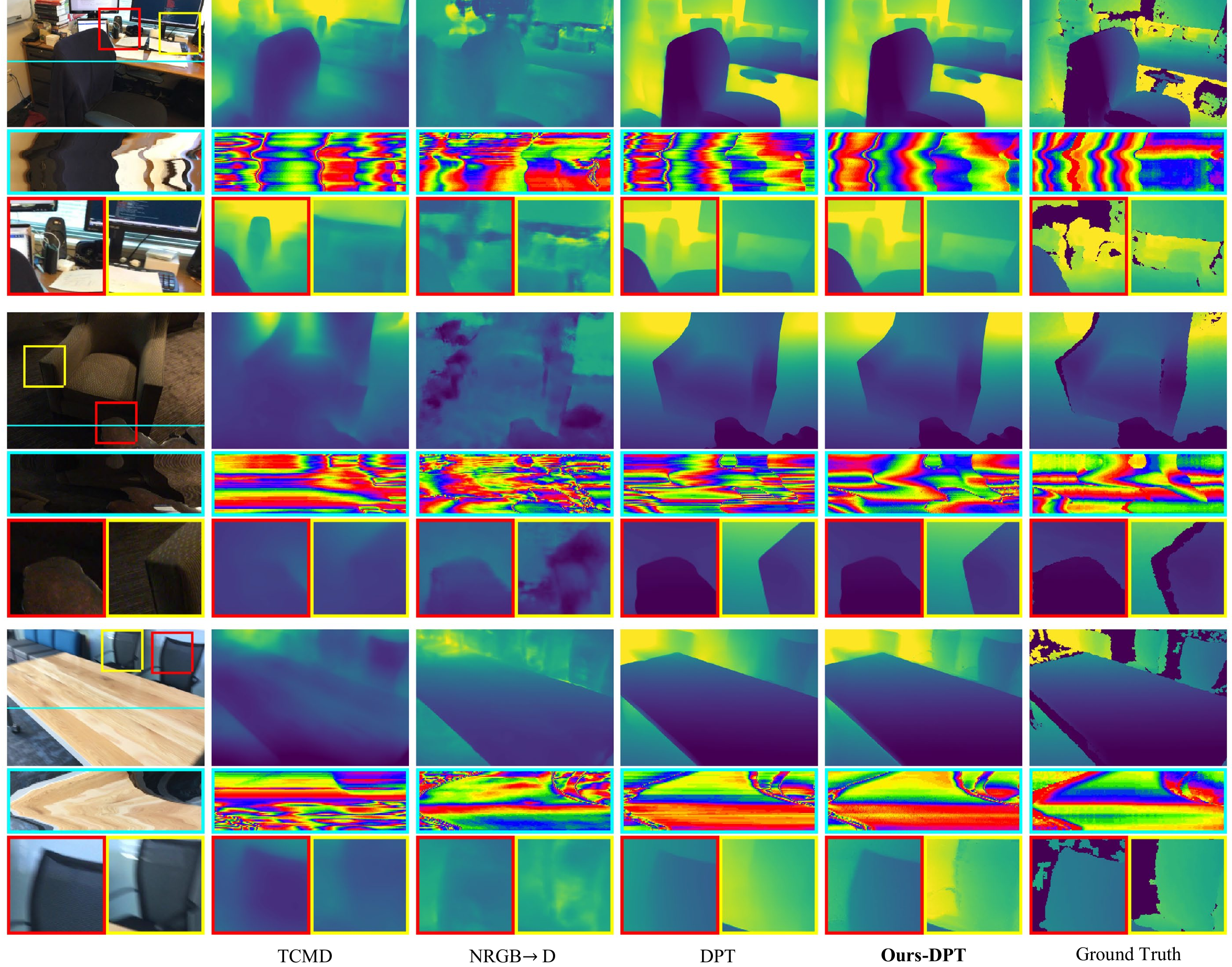}
   \caption{Qualitative comparison of the baseline monocular methods, and our approach with a DPT~\cite{ranftl2021} backbone. We visualize consistency as variation in the \textcolor{cyan}{cyan} slice along the temporal dimension. Our method is more consistent, and captures finer spatial details. }
   \label{fig:results-qualitative-mono}
   \vspace{-3mm}
\end{figure*}

%% file: tables/ablation-pipeline.tex
\begin{table}
\begin{center}
\begin{tabular}{lcccc}
\toprule
 & TCM$\uparrow$ & TCC$\uparrow$ & OPW$\downarrow$ & RAE$\downarrow$\\
\midrule
\small{No Temporal Fusion} &  
\small{-11.4$\%$} & \small{-12.1$\%$} & \small{+27.7$\%$} & \small{+65.7$\%$} \\
\small{No Spatial Fusion} &  \small{-7.44$\%$} & \small{-9.06$\%$} & \small{+39.6$\%$} & \small{+17.1$\%$} \\
\small{No Global PC} & \small{-7.45$\%$} & \small{-9.57$\%$} & \small{+39.5$\%$} & \small{+13.0$\%$} \\
\bottomrule
\end{tabular}
\caption{Ablating our pipeline to evaluate its contribution to the final result. We list the percentage change with respect to \textbf{Ours-DPT} in Table~\ref{table:results-mono} on the dynamic MPI Sintel dataset. For TCM and TCC higher is better, and vice versa for OPW and RAE.}
\label{table:ablation}
\end{center}
\vspace{-5mm}
\end{table}

%% file: tables/temporal-networks-compare.tex
\begin{table}[h!]
\begin{center}
\begin{tabular}{lccc}
\toprule
 & Params. (M) & GMACs & Runtime (ms) \\
\midrule
\small{RAFT (20 iters.)} & 5.26 & 309.5 & 86.2\\
\small{Mask-RCNN} & 44.2 & 134.6 & 42.4\\
\small{Our $\Theta(\cdot)$} & 4.45 & 35.04 & 7.20\\
\small{Our $\Phi(\cdot)$} & 4.44 & 37.14 & 5.10\\
\bottomrule
\end{tabular}
\caption{Run-time and complexity evaluation of our two networks $\Theta(\cdot)$ and $\Phi(\cdot)$ with a 512$\times$512 image on an NVIDIA GeForce RTX 3080 GPU. We use the default 20 iterations for RAFT.}
\label{table:filtering-networks-comparison}
\end{center}
\vspace{-8mm}
\end{table}

%% file: src/5.conclusion.tex
\section{Conclusion}
\label{sec:conclusion}
We presented a method for temporally consistent video depth estimation in online settings. Our approach uses a global point cloud, and learnt image-space fusion to encourage consistency and preserve accuracy. 
While we assumed camera poses and the scale of monocular depth was known, these can also be computed using a point cloud alignment algorithm such as Iterative Closest Point (ICP)~\cite{izadi2011, newcombe2011}, or any visual SLAM system~\cite{fuentes2015}. In the interest of generality, our method does not assume any particular depth-estimation algorithm. But the results may be improved by conditioning the estimation of $d^t$ on the reprojected depth $\prior{d}$. Furthermore, while our approach does not utilize the second view in stereo settings, it may be possible to exploit left-right view consistency in addition to temporal consistency~\cite{cochran1992}. We leave these directions for future work.

%% file: src/A1.results.tex
\input{tables/results-cvd-extended}

\section{Video Results}
\label{sec:video-results}
We encourage the reader to view the results in the supplemental video on our project webpage. We compare our approach qualitatively to SOTA monocular and stereo depth estimation methods on scenes from the MPI Sintel~\cite{butler2012, wulff2012} and ScanNet~\cite{dai2017} datasets.

\section{Extended Results for CVD~\cite{li2019}}
Further evaluation of our method and CVD on the MPI Sintel dataset is presented in Fig.~\ref{fig:results-qualitative-cvd} and Table~\ref{table:results-cvd-extended}. As CVD uses the Mannequin Challenge (MC)~\cite{li2019} network, we use it as the backbone for our method too for fair evaluation. CVD runs offline, fine-tuning the backbone network for $\sim$20 minutes at inference-time. Consequently, its performance is state-of-the-art and represents the high bar for temporal consistency. Our method achieves a good portion of the gains of CVD while being generalizable and online. Furthermore, the constraints for optimizing CVD are obtained from COLMAP~\cite{schoenberger2016mvs, schoenberger2016sfm} for static regions of the scene only. As a result CVD can sometimes filter moving objects as seen in the first row of Fig.~\ref{fig:results-qualitative-cvd}.

\section{Blender Dataset}
\label{sec:blender-dataset}
\begin{figure}[h]
  \centering
   \includegraphics[width=1.0\linewidth, trim={0.0cm, 0.0cm, 4.0cm, 0.0cm}, clip]{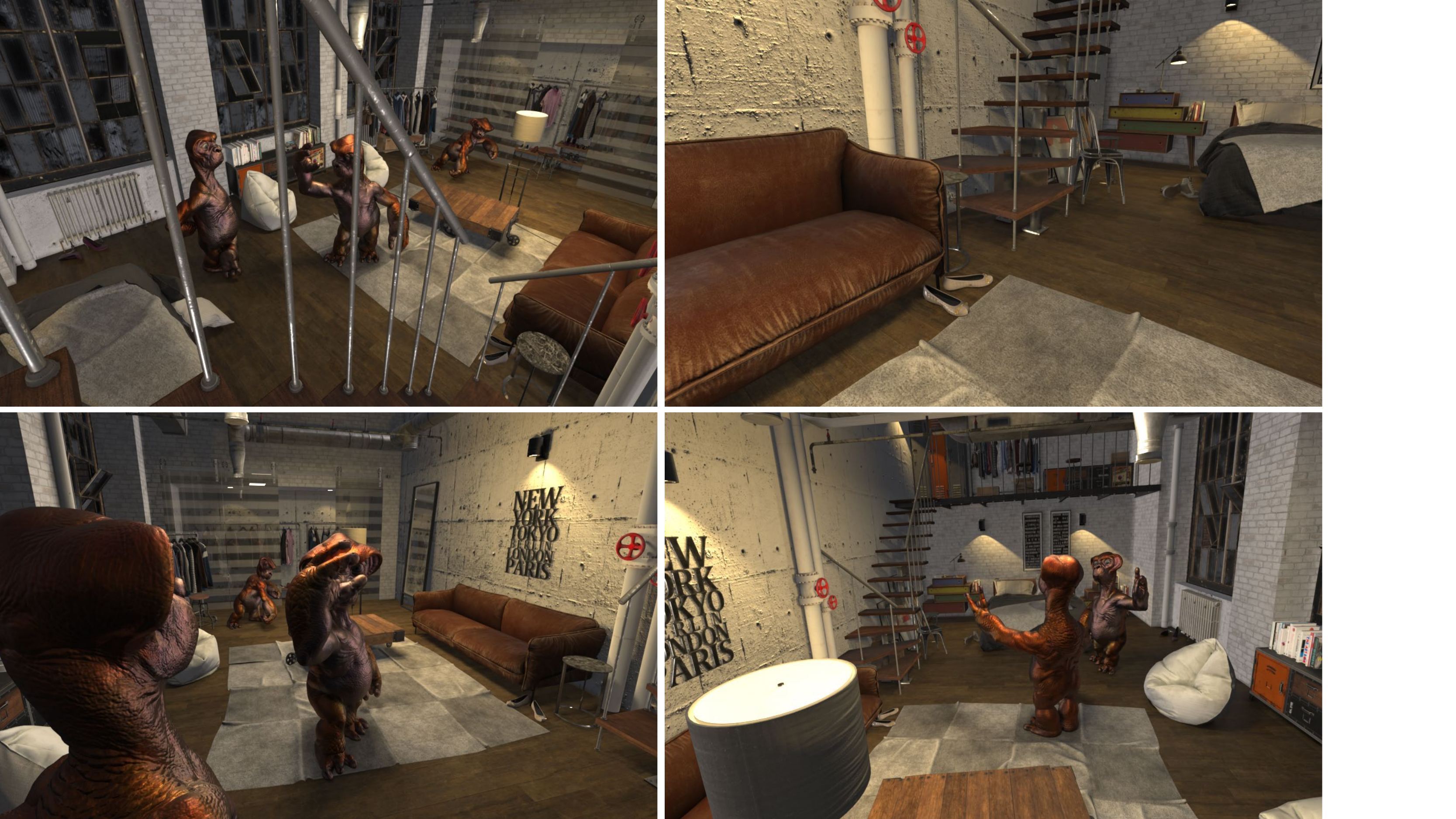}
   \caption{Samples from the custom Blender dataset we rendered to train the temporal fusion network. It consists of 50 60-frame sequences with human-like motion in an indoor setting. We augment the dataset using the techniques described in Sec.~\ref{sec:data-augmentation}}.
   \label{fig:blender-dataset}
\end{figure}

Figure~\ref{fig:blender-dataset} shows samples from the custom Blender dataset we use for fine-tuning the temporal fusion network. The scene consists of organically moving and deforming humanoids in an indoor setting. We generate RGB images, ground truth depth, and camera poses for 50 stereo sequences of 60 frames each ($50\times2\times60=6000$~frames at 1280$\times$720 pixels). The camera poses are manually initialized but follow random trajectories. We use both left and right cameras to generate samples and use data augmentation to increase the diversity of the training data. 

\section{Data Augmentation}
\label{sec:data-augmentation}
We augment both the Blender dataset, and the FlyingThings3D~\cite{mayer2016} dataset described in Sec.~4 of the paper by applying random perturbations to the hue, saturation, brightness and contrast of the RGB images. We also randomly add camera shot noise to simulate real-world capture settings. Further, we generate training samples as random crops of 500$\times$500 and 320$\times$320 pixels respectively from the FlyingThings3D and Blender dataset. We encourage both networks to learn a depth scale invariant representation by randomly scaling the input depth maps by $s \in [0.2, 2]$.

\input{figures/results-epi-cvd}

%% file: tables/results-cvd-extended.tex
\setlength{\tabcolsep}{5pt}

\begin{table*}
\begin{center}
\begin{tabular}{lcccccccccccc}
\toprule
\multirow{2}{*}{} & \multicolumn{12}{c}{MPI Sintel Dataset} \\
\midrule
 & OPW$\downarrow$ & SC$\downarrow$ & RTC$\uparrow$ & TCM$\uparrow$ & TCC$\uparrow$ &  SD(L1)$\downarrow$ & RAE$\downarrow$ & RMS$\downarrow$ & RMS(log)$\downarrow$ & $\delta_1\uparrow$ & $\delta_2\uparrow$ & $\delta_3\uparrow$\\

\midrule
\small{MC~\cite{li2019}} & 0.553 &	0.608 & 0.233 & 0.423 & 0.402 & 0.619 & 0.371 &	3.771 &	0.517 & 0.462 &	0.690 &	0.820  \\
\midrule
\small{CVD-MC~\cite{luo2020}} &
\makecell{{0.226} \\[-0.1cm] \footnotesize{\color{mygreen}{-59.1$\%$}}} &	
\makecell{{0.245} \\[-0.1cm] \footnotesize{\color{mygreen}{-59.7$\%$}}} &
\makecell{{0.402} \\[-0.1cm] \footnotesize{\color{mygreen}{+72.5$\%$}}} & 
\makecell{{0.515} \\[-0.1cm] \footnotesize{\color{mygreen}{+21.7$\%$}}} & 
\makecell{{0.460} \\[-0.1cm] \footnotesize{\color{mygreen}{+14.4$\%$}}} &
\makecell{0.589 \\[-0.1cm] \footnotesize{\color{mygreen}{-4.85$\%$}}} & 
\makecell{{0.338} \\[-0.1cm] \footnotesize{\color{mygreen}{-8.89$\%$}}} &	
\makecell{{3.078} \\[-0.1cm] \footnotesize{\color{mygreen}{-18.4$\%$}}} &	
\makecell{{0.494} \\[-0.1cm] \footnotesize{\color{mygreen}{-4.66$\%$}}} &
\makecell{{0.493} \\[-0.1cm] \footnotesize{\color{mygreen}{+6.71$\%$}}}& 
\makecell{{0.702} \\[-0.1cm] \footnotesize{\color{mygreen}{+1.74$\%$}}} & 
\makecell{0.819 \\[-0.1cm] \footnotesize{\color{red}{-0.12$\%$}}} \\

\small{Ours-MC} &
\makecell{0.377 \\[-0.1cm] \footnotesize{\color{mygreen}{-31.8$\%$}}} &	
\makecell{0.399 \\[-0.1cm] \footnotesize{\color{mygreen}{-34.4$\%$}}} &	
\makecell{0.360 \\[-0.1cm] \footnotesize{\color{mygreen}{+54.5$\%$}}} & 
\makecell{0.443 \\[-0.1cm] \footnotesize{\color{mygreen}{+4.73$\%$}}} & 
\makecell{0.423 \\[-0.1cm] \footnotesize{\color{mygreen}{+5.22$\%$}}} &
\makecell{{0.583} \\[-0.1cm] \footnotesize{\color{mygreen}{-5.82$\%$}}} & 
\makecell{0.361 \\[-0.1cm] \footnotesize{\color{mygreen}{-2.70$\%$}}} &	
\makecell{3.711 \\[-0.1cm] \footnotesize{\color{mygreen}{-1.59$\%$}}} &	
\makecell{0.506 \\[-0.1cm] \footnotesize{\color{mygreen}{-2.13$\%$}}} &	
\makecell{0.466 \\[-0.1cm] \footnotesize{\color{mygreen}{+0.87$\%$}}}& 
\makecell{0.696 \\[-0.1cm] \footnotesize{\color{mygreen}{+0.87$\%$}}} & 
\makecell{{0.823} \\[-0.1cm] \footnotesize{\color{mygreen}{+0.37$\%$}}} \\

\bottomrule
\end{tabular}
\caption{Comparing our approach to CVD using the Mannequin Challenge~(MC) backbone for both methods. The percentage change in each metric over the baseline (MC) is listed in color. CVD fine-tunes the backbone for each scene ($\sim$20 mins) and runs offline.}
\label{table:results-cvd-extended}
\end{center}
\end{table*}

%% file: figures/results-epi-cvd.tex
\begin{figure*}[h]
  \centering
   \includegraphics[width=0.90\linewidth, trim={14.0cm, 0.0cm, 0.0cm, 0.0cm}, clip]{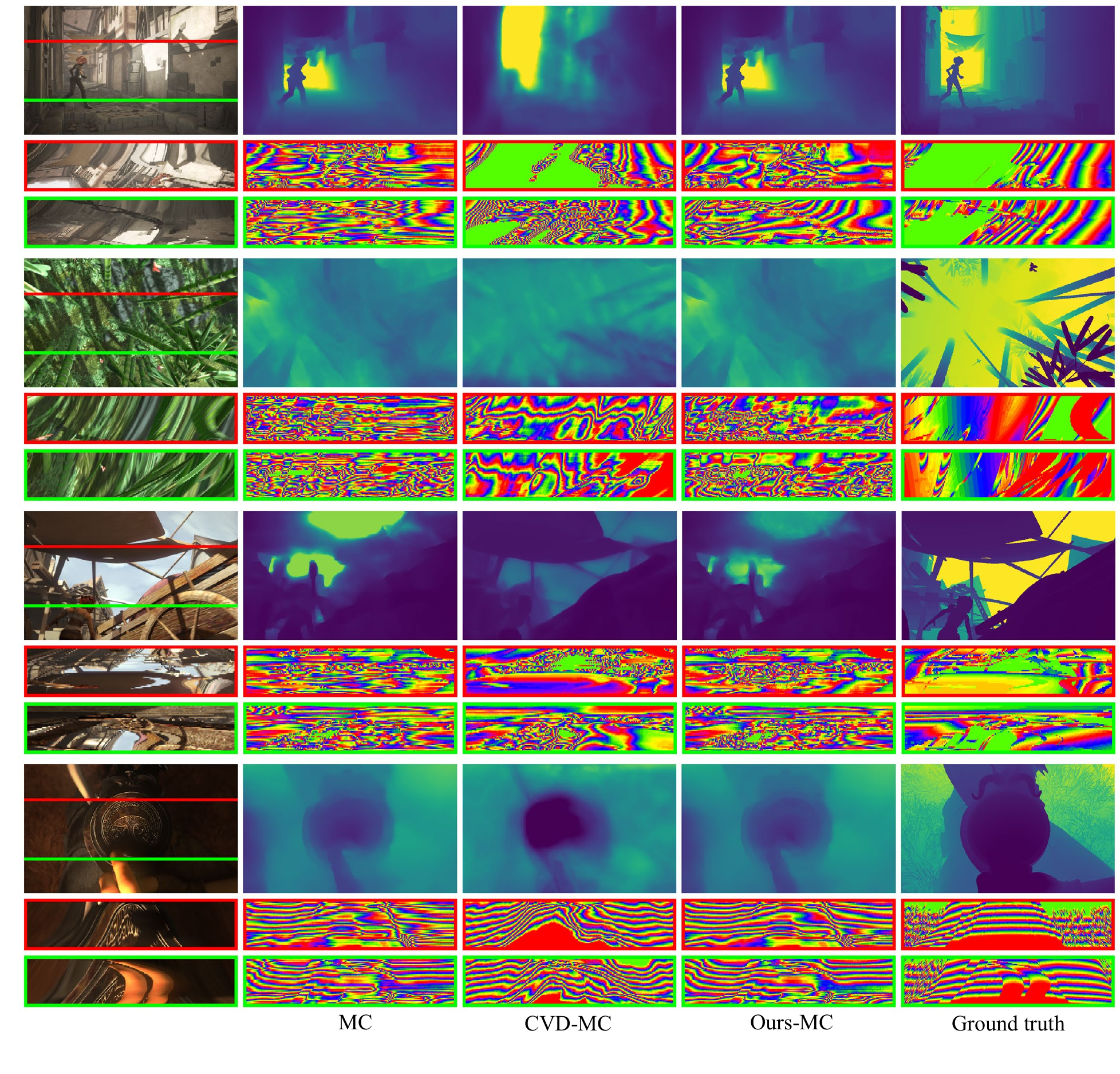}
   \caption{Qualitative comparison of our method with CVD~\cite{luo2020} on the MPI Sintel dataset. We use an MC~\cite{li2019} backbone for both methods. The \textcolor{green}{green} and \textcolor{red}{red} boxes show an angular slice along the temporal dimension.}
   \label{fig:results-qualitative-cvd}
\end{figure*}

%% file: src/A2.architecture.tex
\section{Network Architectures}
\label{sec:network-architectures}
The detailed architectural specifications of the temporal and spatial fusion networks are provided in Tables~\ref{table:network-arch-theta} and~\ref{table:network-arch-phi} respectively. Both networks are based on a four-layer U-Net architecture, with the temporal fusion network having two additional sets of three residual blocks for pre-processing the color and depth branches.

\begin{table}[h!]
\begin{center}
\begin{tabular}{lccccc}
\toprule
\multicolumn{6}{c}{Temporal Fusion Network} \\
\midrule
Layer & \textbf{k} & \textbf{s} & \textbf{p} & Channels & Input \\
\midrule
down1-1 & & & & 2/2 & $\obv{d} \oplus \prior{d}$ \\
down1-2 & & & & 6/6 & $\obv{c} \oplus \prior{c}$ \\
resblock2-1 & 5 & 1 & 2 & 2/8 & down1-1 \\
resblock2-2 & 3 & 1 & 1 & 8/16 & resblock2-1 \\
resblock2-3 & 3 & 1 & 1 & 16/24 & resblock2-2 \\
resblock2-4 & 5 & 1 & 2 & 6/8 & down1-2 \\
resblock2-5 & 3 & 1 & 1 & 8/16 & resblock2-4 \\
resblock2-6 & 3 & 1 & 1 & 16/24 & resblock2-5 \\
up3-1 & & & & 24/24 & resblock2-3 \\
up3-2 & & & & 24/24 & resblock2-6  \vspace{1mm} \\
\midrule
conv1-1 & 3 & 1 & 1 & 52/24 & 
\makecell{up3-1 $\oplus$
up3-2~$\oplus$ \\ $\obv{d}\,\oplus\,\obv{c}$} \vspace{1mm} \\
conv1-2 & 3 & 1 & 1 & 24/24 & conv1-1 \\
maxpool1-1 & 3 & 1 & 1 & 24/24 & conv1-2 \\

conv2-1 & 3 & 1 & 1 & 24/48 & maxpool1-1 \\
conv2-2 & 3 & 1 & 1 & 48/48 & conv2-1 \\
maxpool2-1 & 3 & 1 & 1 & 48/48 & conv2-2 \\

conv3-1 & 3 & 1 & 1 & 48/96 & maxpool2-1 \\
conv3-2 & 3 & 1 & 1 & 96/96 & conv3-1 \\
maxpool3-1 & 3 & 1 & 1 & 96/96 & conv3-2 \\

conv4-1 & 3 & 1 & 1 & 96/192 & maxpool3-1 \\
conv4-2 & 3 & 1 & 1 & 192/192 & conv4-1 \\
maxpool4-1 & 3 & 1 & 1 & 192/192 & conv4-2 \\

conv5-1 & 3 & 1 & 1 & 192/384 & maxpool4-1 \\
conv5-2 & 3 & 1 & 1 & 384/384 & conv5-1 \\

up6-1 & & & & 384/384 & conv5-2 \\
conv6-2 & 3 & 1 & 1 & 576/192 & conv4-2 $\oplus$ up6-1 \\
conv6-3 & 3 & 1 & 1 & 192/192 & conv6-2 \\

up7-1 & & & & 192/192 & conv6-3 \\
conv7-2 & 3 & 1 & 1 & 288/96 & conv3-2 $\oplus$ up7-1 \\
conv7-3 & 3 & 1 & 1 & 96/96 & conv7-2 \\

up8-1 & & & & 96/96 & conv7-3 \\
conv8-2 & 3 & 1 & 1 & 144/48 & conv2-2 $\oplus$ up8-1 \\
conv8-3 & 3 & 1 & 1 & 48/48 & conv8-2 \\

up9-1 & & & & 48/48 & conv8-3 \\
conv9-2 & 3 & 1 & 1 & 72/24 & conv1-2 $\oplus$ up9-1 \\
conv9-3 & 3 & 1 & 1 & 24/1 & conv9-2 \\
\bottomrule
\end{tabular}
\caption{The network architecture for the temporal fusion network $\Theta(\cdot)$~(Sec.~3.1, main paper); \textbf{k},~\textbf{s},~\textbf{p} denote the kernel size, stride and padding respectively, $\oplus$ is concatenation along the channel dimension, and ``down''/~``up'' is $\times$2 bilinear down/up-sampling.  All convolutional layers in the lower section except \textbf{conv9-3} are followed by a ReLU activation and instance norm; \textbf{conv9-3} has a sigmoid activation. All convolutions are 2D with a dilation of one.}
\label{table:network-arch-theta}
\end{center}
\vspace{-6mm}
\end{table}

\begin{table}[h!]
\begin{center}
\begin{tabular}{lccccc}
\toprule
\multicolumn{6}{c}{Spatial Fusion Network} \\
\midrule
Layer & \textbf{k} & \textbf{s} & \textbf{p} & Channels & Input \\
\midrule
conv1-1 & 3 & 1 & 1 & 4/24 & $\obv{c} \oplus \obv{d}$ OR  $\obv{c} \oplus \dtemp$\\
conv1-2 & 3 & 1 & 1 & 24/24 & conv1-1 \\
maxpool1-1 & 3 & 1 & 1 & 24/24 & conv1-2 \\

conv2-1 & 3 & 1 & 1 & 24/48 & maxpool1-1 \\
conv2-2 & 3 & 1 & 1 & 48/48 & conv2-1 \\
maxpool2-1 & 3 & 1 & 1 & 48/48 & conv2-2 \\

conv3-1 & 3 & 1 & 1 & 48/96 & maxpool2-1 \\
conv3-2 & 3 & 1 & 1 & 96/96 & conv3-1 \\
maxpool3-1 & 3 & 1 & 1 & 96/96 & conv3-2 \\

conv4-1 & 3 & 1 & 1 & 96/192 & maxpool3-1 \\
conv4-2 & 3 & 1 & 1 & 192/192 & conv4-1 \\
maxpool4-1 & 3 & 1 & 1 & 192/192 & conv4-2 \\

conv5-1 & 3 & 1 & 1 & 192/384 & maxpool4-1 \\
conv5-2 & 3 & 1 & 1 & 384/384 & conv5-1 \\

up6-1 & & & & 384/384 & conv5-2 \\
conv6-2 & 3 & 1 & 1 & 576/192 & conv4-2 $\oplus$ up6-1 \\
conv6-3 & 3 & 1 & 1 & 192/192 & conv6-2 \\

up7-1 & & & & 192/192 & conv6-3 \\
conv7-2 & 3 & 1 & 1 & 288/96 & conv3-2 $\oplus$ up7-1 \\
conv7-3 & 3 & 1 & 1 & 96/96 & conv7-2 \\

up8-1 & & & & 96/96 & conv7-3 \\
conv8-2 & 3 & 1 & 1 & 144/48 & conv2-2 $\oplus$ up8-1 \\
conv8-3 & 3 & 1 & 1 & 48/48 & conv8-2 \\

up9-1 & & & & 48/48 & conv8-3 \\
conv9-2 & 3 & 1 & 1 & 72/48 & conv1-2 $\oplus$ up9-1 \\
conv9-3 & 3 & 1 & 1 & 48/24 & conv9-2 \\
conv9-4 & 3 & 1 & 1 & 24/1 & conv9-3 \\

\bottomrule
\end{tabular}
\caption{The network architecture for the spatial fusion network $\Phi(\cdot)$~(Sec.~3.2, main paper); \textbf{k},~\textbf{s},~\textbf{p} denote the kernel size, stride and padding respectively, $\oplus$ is concatenation along the channel dimension, and ``up'' is $\times$2 bilinear up-sampling. }
\label{table:network-arch-phi}
\end{center}
\end{table}